\documentclass{bmvc2k}
\usepackage{adjustbox}
\usepackage[capitalise]{cleveref}
\usepackage{booktabs}
\usepackage{wrapfig}
\usepackage{floatflt}
\usepackage{capt-of}

\title{From Open Vocabulary to Open World: Teaching Vision Language Models to Detect Novel Objects}

\addauthor{Zizhao Li}{zizhao.li1@student.unimelb.edu.au}{1}
\addauthor{Zhengkang Xiang}{zhengkangx@student.unimelb.edu.au}{1}
\addauthor{Joseph West}{joseph.west@unimelb.edu.au}{1}
\addauthor{Kourosh Khoshelham}{k.khoshelham@unimelb.edu.au}{1}

\addinstitution{
 The University of Melbourne\\
 Parkville, VIC, Australia
}

\runninghead{Li, Xiang, West, Khoshelham}{From Open Vocabulary to Open World}

\def\etal{\emph{et al}\bmvaOneDot}

\begin{document}

\maketitle

\begin{abstract}
Traditional object detection methods operate under the closed-set assumption, where models can only detect a fixed number of objects predefined in the training set. Recent works on open vocabulary object detection (OVD) enable the detection of objects defined by an in-principle unbounded vocabulary, which reduces the cost of training models for specific tasks. However, OVD heavily relies on accurate prompts provided by an ``oracle'', which limits their use in critical applications such as driving scene perception. OVD models tend to misclassify near-out-of-distribution (NOOD) objects that have similar features to known classes, and ignore far-out-of-distribution (FOOD) objects.
To address these limitations, we propose a framework that enables OVD models to operate in open world settings, by identifying and incrementally learning previously unseen objects.
To detect FOOD objects, we propose Open World Embedding Learning (OWEL) and introduce the concept of Pseudo Unknown Embedding which infers the location of unknown classes in a continuous semantic space based on the information of known classes. 
We also propose Multi-Scale Contrastive Anchor Learning (MSCAL), which enables the identification of misclassified unknown objects by promoting the intra-class consistency of object embeddings at different scales. 
The proposed method achieves state-of-the-art performance on standard open world object detection and autonomous driving benchmarks while maintaining its open vocabulary object detection capability.
The code is available at \href{https://github.com/343gltysprk/ovow}{https://github.com/343gltysprk/ovow}.
\end{abstract}

\section{Introduction}
\label{sec:intro}
Object detection is a fundamental computer vision task, which involves localization and classification of foreground objects. Although there has been significant progress in this area, many methods rely on the closed-set assumption~\cite{rcnn,fasterrcnn,fastrcnn,yolo,ssd,retinanet,tian2019fcos,detr}, where all object categories to be predicted are available in the training set. 

In many real world applications, such as autonomous driving, the closed-set assumption is unrealistic and even dangerous, because it forces the model to misclassify or ignore unknown objects~\cite{elephant}. Scheirer \etal \cite{tosr} defines the problem of rejecting unknown classes ($\mathcal{U}$) and simultaneously classifying known classes ($\mathcal{K}$) as open set recognition (OSR). Subsequent works \cite{miller18,elephant} extend this problem to open set object detection (OSOD). Joseph \etal~\cite{TowardsOWOD} further proposes Open World Object Detection (OWOD), which involves detecting both known and unknown objects and incrementally learning new classes.


Open world object detection is a challenging task due to the complexity of both open-set recognition \cite{miller18,elephant} and incremental learning \cite{incremental}. The model must generalize beyond predefined classes to capture the objectness of diverse unknown objects, and avoid confusing them with known classes. Additionally, it needs to incrementally learn new classes without forgetting previously acquired knowledge. Despite some progress \cite{TowardsOWOD,OW-DETR,ost,UC-OWOD,ALLOW,prob,cat,randbox,sun2024exploring} in this area, several key issues still remain unresolved. Many existing methods 
perform poorly at discovering unknown objects, leading to low recall for unknown classes. Additionally, existing OWOD methods \cite{TowardsOWOD,OW-DETR,sun2024exploring,randbox,prob,ALLOW,ost,UC-OWOD,cat} employ a replay strategy during incremental learning, where data from previous tasks are reintroduced during the training of new classes. This strategy is suboptimal, leading to unnecessary consumption of computing resources and storage.

Recently, the rise of visual-language pre-training has spawned a new area: open vocabulary object detection (OVD). In principle, OVD can detect novel classes defined by an unbounded (open) vocabulary at inference. Naturally, it is able to incrementally learn new classes by adding new prompts. 
With a limited vocabulary the model tends to misclassify near-out-of-distribution (NOOD) objects that have similar semantics to known classes, and ignore far-out-of-distribution (FOOD) objects.
In real-world applications such as autonomous driving, where countless object-types cannot be initially included in the text prompt, OVD will inevitably fail when presented with OOD objects. 


To tackle this challenge, we propose a framework that enables OVD models to operate in open-world settings. To detect FOOD objects and incrementally learn the new classes, we propose Open World Embedding Learning (OWEL). OWEL optimizes parameterized class embeddings rather than fine-tuning the whole model to learn new classes, inherently avoiding catastrophic forgetting. We further introduce the novel concept of Pseudo Unknown Embedding, which constructs a text embedding to detect FOOD objects based on current known classes and generic objectness. 

To detect NOOD objects, we propose Multi-Scale Contrastive Anchor Learning 
(MSCAL). Contrastive learning has been used in OOD detection~\cite{csi,cider,PALM2024} to pull similar samples together and push dissimilar samples apart in the representation space. 
The key insight of MSCAL is that in an open world setting, object classes are introduced gradually. As new classes are introduced, the decision boundaries for known classes shift within the shared feature space.  We formulate the task of unknown object identification in the open-set setting as a series of deep one-class classification~\cite{docc} problems. For each class $i$, we use an individual non-linear projector to map the feature pyramid into a class-specific representation space, contrasting embeddings with class anchors. Positive samples from class $i$, at different scales, maximize their similarity to the class anchor, while embeddings from other classes and the background act as negative samples. MSCAL ensures that embeddings of known classes at different scales are tightly clustered around their corresponding anchors, while unknown object embeddings are left out of the clusters and can be rejected based on distance.

Our method significantly surpasses  state-of-the-art (SOTA) performance in U-Recall on the M-OWODB~\cite{TowardsOWOD} and S-OWODB~\cite{OW-DETR} benchmarks, while maintaining leading performance in other metrics. We further evaluate our method on a novel benchmark based on nuScenes~\cite{nuscenes2019}, where it also achieves the best results. More importantly, our method preserves the zero-shot capability of the OVD model, as our implementation only optimizes text embeddings and additional MSCAL modules, while keeping the OVD model's weights frozen.
Therefore, our method provides a unified framework for both open vocabulary learning and open world learning. 
The contributions of this work are as follows:
\begin{itemize}
    \item We propose a framework that enables OVD models to operate in open world settings by identifying unknown objects and incrementally learning new classes, thereby unifying open vocabulary learning and open world learning within the same framework.
    \item We propose a novel method, Open World Embedding Learning (OWEL), to enable the discovery and incremental learning of new classes without fine-tuning the whole model or requiring exemplars of previous tasks.
    \item We propose Multi-Scale Contrastive Anchor Learning (MSCAL), which reduces the known-unknown confusion in OWOD by clustering known class embeddings at different scales around class-specific anchors. 
    \item We propose a new benchmark for OWOD application in autonomous driving based on the commonly used nuScenes dataset.
\end{itemize}

\section{Related Works}
\label{sec:formatting}

Closed set object detection has been extensively studied over the past decade~\cite{rcnn,fasterrcnn,fastrcnn,yolo,ssd,retinanet,tian2019fcos,detr}. To handle unknown objects and incrementally learn new classes, Joseph \etal~\cite{TowardsOWOD} first proposed an open world object detector, which extends the Faster R-CNN~\cite{fasterrcnn} with contrastive clustering and energy-based unknown-object identification. Subsequent works~\cite{OW-DETR,ALLOW} used contextual information to improve unknown identification and knowledge transfer between known and unknown classes. To effectively detect unknown objects, Zohar \etal~\cite{prob} proposed Probabilistic Objectness (PROB) to estimate the objectness of different proposals. Wang \etal \cite{randbox} introduced random proposals in detector training to encourage the unknown discovery and reduce confusion between known and unknown classes. Ma \etal \cite{cat} proposed decoupling object localization and classification via cascade decoding. Hyp-OW~\cite{doan_2024_HypOW} uses hyperbolic distance to enhance open world learning. Sun \etal \cite{sun2024exploring} further de-correlates objectness and class information by enforcing orthogonality. In some recent works~\cite{skdf,Maaz2022Multimodal,zohar2023open}, foundation models are employed in open-world learning.

While open vocabulary object detection (OVD) aims to detect novel classes with the help of vocabulary knowledge~\cite{ovlsurvey}.  OVD models \cite{gu2022openvocabulary,hierkd,Hanoona2022Bridging,glip,zhang2022glipv2,yolow,wu2023cora,Du_2022_CVPR} typically match the image embeddings with the text embeddings. When presented with an object that is not in the prompt, the model will either assign an incorrect label of the nearest match (NOOD case) or no detection at all (FOOD case). 

Out-of-distribution (OOD) detection~\cite{mcm,maxlogit,energy,vim,cider,odin,md,sun2022knnood,csi,Zolfi_2024_CVPR,mcm,jiang2024negative,clipn,locoop,negativeprompts,npcvpr24} has received significant attention in recent years.
Despite the relevance of OOD detection in this context, there has been limited exploration of integrating OOD detection methods into OVD frameworks.

\section{Method}
\begin{figure*}[ht]
  \centering
    \includegraphics[width=0.9\textwidth]{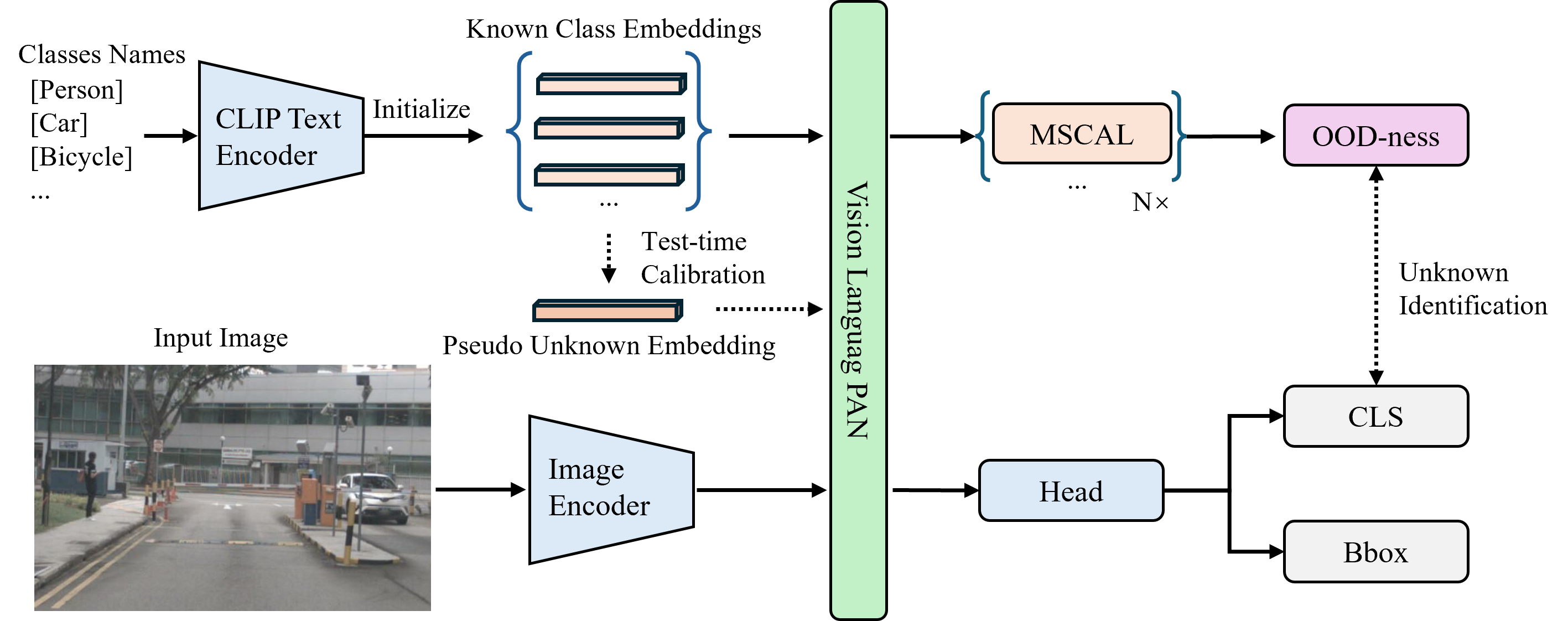}
    \caption{\textbf{Overview of the proposed method.} During model training, we first initialize known class embeddings with a pretrained CLIP text encoder~\cite{clip}. 
    The image encoder extracts a multi-scale feature map from the input.
    Then the RepVL-PAN \cite{yolow} uses multi-level cross-modal fusion to combine image and text features, forming the feature pyramids. The detection head predicts the class label based on image-text similarity and regresses the bounding box. The detection loss is used to update the known class embeddings. Concurrently, MSCAL modules are trained to maximize the similarity between class anchor and spatial locations at different scales, and output a multi-scale score map to indicate whether an embedding is out-of-distribution (OOD) relative to a specified class. During the inference, the OOD map extracted by MSCAL is used to reduce known-unknown confusion. In addition, the pseudo unknown embedding used to discover unknown classes is constructed from the optimized known class embeddings and the generic ``objectness'' semantic concept.}
    \label{fig:main}
    \vspace{-0.5cm}
\end{figure*}

\paragraph{Problem Definition}
Open World Object Detection (OWOD) \cite{TowardsOWOD} aims to detect both known and unknown objects while continuously learning new classes. At an arbitrary stage $t$, we consider the known classes as $\mathcal{K}^t = \{1,...,N\}$, and unknown classes as  $\mathcal{U}$.
An OWOD model should be able to detect objects in $\mathcal{K}^t$ and $\mathcal{U}$, and extend known classes to $\mathcal{K}^{t+1} =\mathcal{K}^{t} \cup \{N+1,...,N+k\}$ when $k$ new classes are incrementally learned. In this way, the object detector continuously discovers and learns new classes in the open world.

\paragraph{General Architecture}
\cref{fig:main} shows the general architecture of the proposed method. Following \cite{yolow}, we use text $T$ and image $I$ as inputs and match the text embeddings with image embeddings to predict class labels and bounding boxes of objects.
Let $W_\mathcal{K}=\{w_1,...,w_N\}$ denote the text embedding of $N$ known classes, which is initialized from class names encoded by the pre-trained CLIP \cite{clip} text encoder. $W_\mathcal{K}$ can be parameterized as model's weight and optimized via Open World Embedding Learning (OWEL).
During inference, a pseudo unknown embedding is constructed and appended to $W_\mathcal{K}$, and the CLIP text encoder is disposable.
The image encoder (DarkNet backbone inherited from Yolo v8~\cite{yolov8_ultralytics,yolov3}) extracts multi-scale features $C$ from the input image $I$. Then the multi-modal neck (RepVL-PAN \cite{yolow}) uses multi-level cross-modal fusion to combine image and text features, forming the feature pyramids $P$. The detection head predicts bounding boxes and class labels by matching the cosine similarity of text embedding with each spatial location in $P$. Concurrently, MSCAL modules make dense predictions of out-of-distribution (OOD) scores and reject unknown bounding boxes in the detection head.
Finally, the redundant predictions are filtered by Non-Maximum Suppression (NMS).

\paragraph{Open World Embedding Learning}
\label{sec:owec}

\begin{figure}[t!]
\begin{minipage}[b]{0.48\linewidth}
  \centering
    \includegraphics[width=0.8\textwidth]{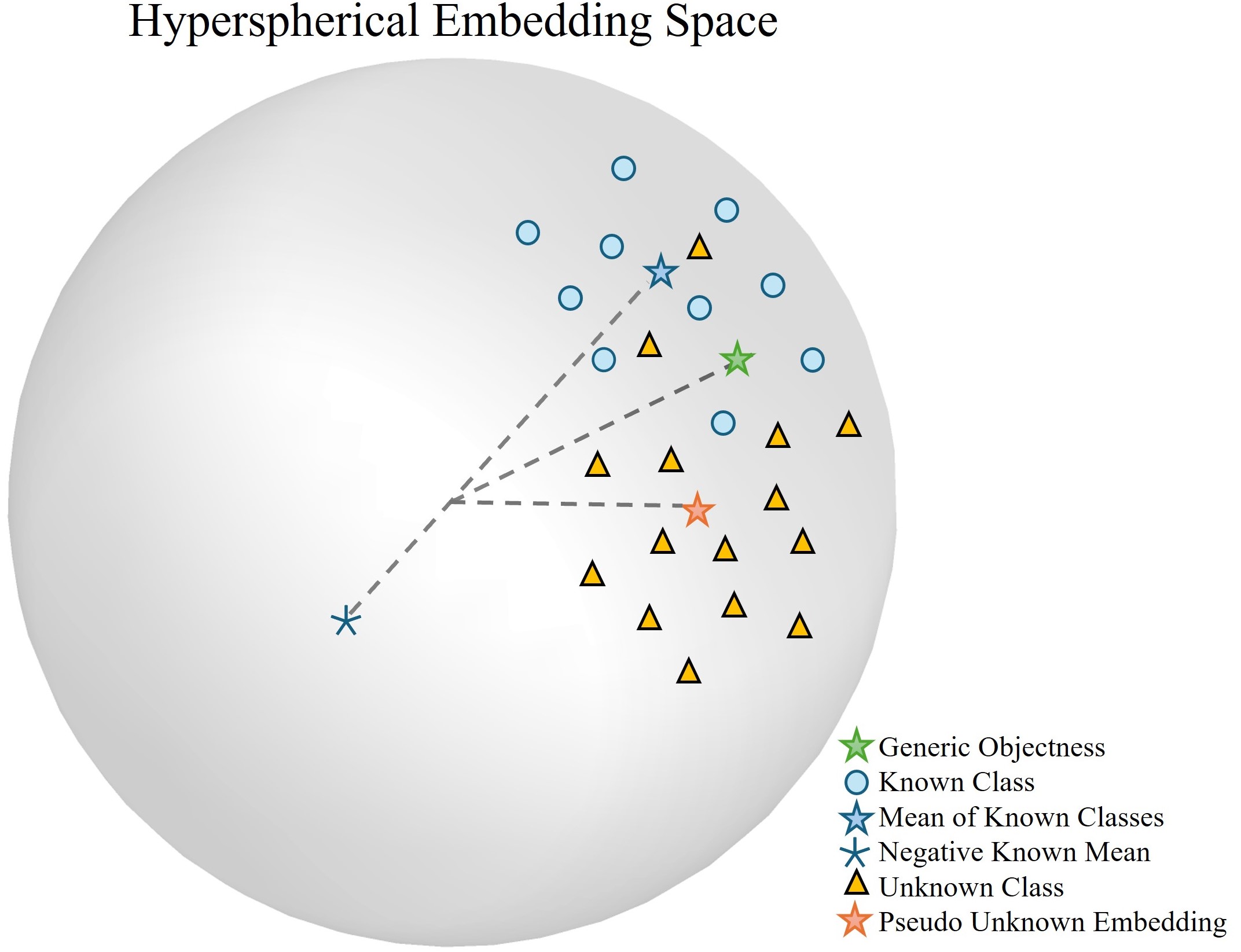}

   \caption{\textbf{Inferring the Pseudo Unknown Embedding in the embedding space.} For CLIP-like models, text embeddings are mapped on a unit hypersphere. 
   }
   \label{fig:owel}
\end{minipage}\hfill
\begin{minipage}[b]{0.48\linewidth}
    \includegraphics[width=\textwidth]{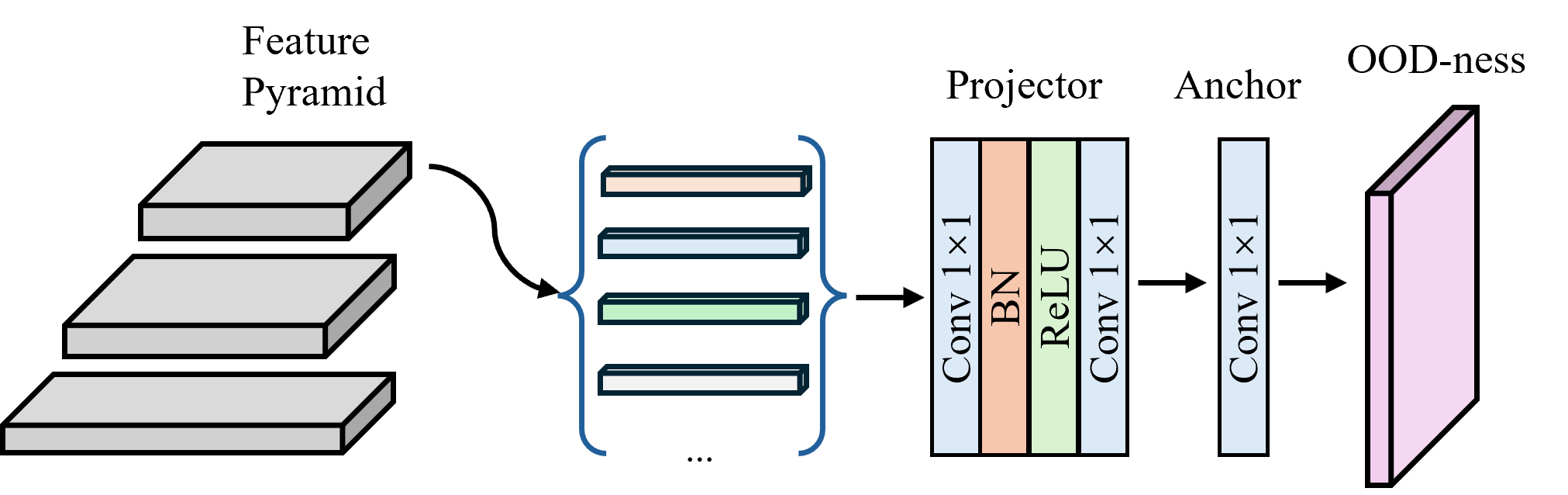}

   \caption{\textbf{MSCAL module.} For each layer in the feature pyramid, all spatial locations will be mapped to a new space and contrasted with class anchors. The design of the projector follows \cite{Wang_2021_ICCV}.
   During inference, their inner product with the class anchor serves as the OOD score.}
   \label{fig:mscal}
\end{minipage}
\vspace{-0.2cm}
\end{figure}

Large pre-trained vision-language models have enabled many prompt learning methods \cite{zhou2022coop,zhou2022cocoop,eco}, which optimize text prompts rather than fine-tune the entire model to improve performance in downstream tasks.  But they are not suitable for OWOD, because known $\mathcal{K}$ and unknown $\mathcal{U}$ classes are ever-changing.
To address this, we propose a simple and effective way to learn new classes and detect unknown objects, called Open World Embedding Learning (OWEL). For $N$ known classes, we initialize known class embeddings $W_\mathcal{K}=\{w_1,...,w_N\}$, and optimize them with the object detection loss. When $k$ new classes are introduced, we freeze $W_\mathcal{K}$ and train new class embeddings.

When an OVD model is given an unknown object, it will either misclassify it as a semantically similar text prompt, or ignore it if the semantic difference is large.
To avoid misclassification, we can model the distribution of known classes to reject out-of-distribution samples. To discover novel objects, we construct a Pseudo Class Embedding $w_{\mathcal{U}}$ representing unknown classes.

Vision-language pretraining aligns the visual embedding space and the textual embedding space, such that the text embedding is equivalent to the corresponding image embedding \cite{clip,zhou2022coop,clipn,vlp}. It has been shown that the relation between text embeddings can be derived through vector offsets in continuous space language models \cite{mikolov-etal-2013-linguistic,Mikolov2013EfficientEO}. As shown in \cref{fig:owel}, let $\overline{w}$ denote the mean text embedding of all known classes, defined as:
\begin{equation}
    \overline{w} =  \sum_{i=1}^{N}\frac{w_i}{N||w_i||}.
\end{equation}


Let $w_0$ denote the text embedding representing generic objectness. Naturally, the text prompt of $w_0$ should be some general words, such as ``object'', which is supported by a previous work~\cite{Maaz2022Multimodal} and our observation on YOLO-World.
The semantics of $w_0$ overlaps with known classes to some extent.
To shift its focus to unknown classes rather than making duplicate detections on known classes, we propose to construct a Pseudo Unknown Embedding, which is specialized to detect far-out-of-distribution (FOOD) objects. 
The Pseudo Unknown Embedding $w_{\mathcal{U}}$ is defined by subtracting the mean of known classes from $w_0$:
\begin{equation}
    w_{\mathcal{U}} = w_0 - \alpha  \frac{\overline{w}}{||\overline{w}||},
\end{equation}

where $\alpha$ is a weight parameter. Since $w_{\mathcal{U}}$ is defined at test time, it is able to dynamically shift its focus when new known classes are added.


\paragraph{Multi-Scale Contrastive Anchor Learning}
\label{sec:mscal}

To effectively identify near-out-of-distribution objects and continuously accommodate new classes, we propose a method called Multi-Scale Contrastive Anchor Learning (MSCAL). Assume we currently have $N$ known classes. For each class $i$, we train a MSCAL module to identify whether a spatial location in feature pyramid $P$ belongs to that class, by maximizing the inner product between the class anchor $\mu_i$ with spatial locations from class $i$, and minimizing the inner product with spatial locations from other classes.

In OWOD, object classes are introduced gradually. New classes will shift the decision boundary of existing classes. Alternatively, we can use a class-specified module to enforce the consistency within this class, and reject all samples from other classes. For each class $i$, we use the individual non-linear projector to map the feature pyramid into a class-specific representation space, where contrastive learning takes place. \cref{fig:mscal} shows the structure of MSCAL module. 
For each class $i$, the non-linear contrastive projector $g_i(\cdot)$ map $P$ to multi-scale feature map $\mathcal{Z}_i$ in lower dimensional space. We view $\mathcal{Z}_i$ as the collection of spatial locations corresponding to class anchor $i$ in the mini-batch, and assume there are $p$ layers of feature maps extracted by projector $g_i(\cdot)$, so the MSCAL loss for class $i$ is 


\begin{equation}
\mathcal{L}_{i}^{con} = \frac{-1}{\left |  \mathcal{Z}^+_i  \right|} \sum_{j=1}^{p} \sum_{z_k \in \mathcal{Z}^+_{ij} }^{} \log
{\frac{\exp{\left(\mu_{ij} \cdot{z}_k/\tau\right)}}{\sum_{m=1}^{p}\sum_{z_n\in \mathcal{Z}_{im}}\exp{\left(\mu_{im} \cdot{z}_n/\tau\right)}}},
\end{equation}

where $\mathcal{Z}^+_i$ denotes the collection of positive samples from class $i$, $\left |  \mathcal{Z}^+_i  \right|$ is the number of positive samples, $\mathcal{Z}^+_{ij}$ is the feature map at layer $j$, the $\cdot$ symbol denotes the inner product, and $\tau$ is a temperature scaling parameter. Accordingly, the loss for all known classes is defined as:
\begin{equation}
\mathcal{L}^{con} = \frac{1}{N} \sum_{i=1}^{N} \mathcal{L}_{i}^{con}.
\end{equation}

MSCAL contrasts class anchors with positive and negative samples across different scales and images.
During the inference, the OOD score at spatial location $z$ is,
$\mathcal{S}(z) = -\max_i \mu_{i} \cdot z$.
The idea is that during training $\mu_{i} \cdot z$ will be maximized if $z$ belongs to class $i$, and minimized otherwise. For an unknown object, $\mathcal{S}(z)$ will be larger than that for known objects. 
We use this OOD score to identify unknown objects proposed by the detection head. 


\paragraph{Incremental Learning}
Our method does not require exemplar replay to prevent catastrophic forgetting. We simply freeze the parameterized text embeddings and MSCAL modules for previously known classes and train new modules with currently known classes.

\paragraph{Model Inference}
Following YOLO-World \cite{yolow}, we match feature maps with text embeddings (including known classes and the Pseudo Unknown Embedding $w_{\mathcal{U}}$) and use OOD scores from the MSCAL module to identify misclassified unknown objects. A region is considered background if it fails to match any text embedding. If it matches a known-class embedding and has a low OOD score, it is classified as a known object. In contrast, a region is identified as an unknown object if it either matches the pseudo unknown embedding $w_{\mathcal{U}}$ or exhibits a high OOD score with respect to the known classes.

\section{Experiments}

\subsection{Datasets}
 We evaluate our method with common open world object detection benchmarks used in previous works \cite{prob,cat,sun2024exploring}, and propose a novel benchmark of OWOD for autonomous driving. Common OWOD benchmarks include the superclass-mixed benchmark (M-OWODB) \cite{TowardsOWOD} and the superclass-separated benchmark (S-OWODB) \cite{OW-DETR}. The M-OWODB benchmark combines COCO \cite{coco} and PASCAL VOC \cite{voc}, while the S-OWODB benchmark is based solely on COCO. Both are divided into four distinct tasks, where the model learns some new classes in each task, while the remaining classes are unknown. 
Additionally, we propose a challenging OWOD benchmark (nu-OWODB) based on nuScenes~\cite{nuscenes2019}, which consists of real-world driving scenes. The nuScenes dataset captures diverse urban environments, including crowded city streets with many dynamic objects, challenging weather conditions, and dense traffic scenarios with occlusions and complex interactions between agents.
In addition, the dataset contains a significant class imbalance, with some classes like cars being much more frequent than others like ambulances or construction vehicles. The nu-OWODB benchmark is divided into three subtasks. Initially, the model is introduced to different types of vehicles. In subsequent tasks, various pedestrians and other traffic participants are introduced. This benchmark simulates the challenges of OWOD in real-world applications.
For the open vocabulary evaluation, we adopt the LVIS minival~\cite{goldg} benchmark, which is widely used in previous works~\cite{DetCLIP,groundingdino,yolow}.

\subsection{Evaluation Metrics}
We evaluate the performance of the proposed model on both known and unknown classes in each task. For known classes, the commonly used metric is mean average precision (mAP). Specifically, the evaluation is further divided into the mAP of previously known classes and currently known classes.
For unknown classes, the primary metric will be unknown class recall (U-Recall), which assesses the model's ability to detect unknown objects. Additionally, we employ wilderness impact (WI) \cite{elephant} and absolute open-set error (A-OSE) \cite{miller18} to measure the extent of the model's confusion between known and unknown classes.

\subsection{Results}

\paragraph{Quantitative Results on Traditional OWOD Benchmarks}
\Cref{table1} shows the OWOD performance on commonly used benchmarks, comparing our method with existing OWOD methods and the unmodified YOLO-World~\cite{yolow}.
Our method significantly outperforms both traditional open-world object detection methods~\cite{TowardsOWOD,OW-DETR,cat,sun2024exploring,ALLOW,UC-OWOD,randbox,ost} based on ImageNet~\cite{imagenet} pretrained backbones, as well as OWOD methods~\cite{Maaz2022Multimodal,skdf} that leverage large-scale vision-language pretraining, in terms of mean average precision (mAP) for known classes and unknown class recall (U-Recall).
Our method also better reduces unknown object confusion, achieving lower WI and A-OSE compared to state-of-the-art methods (shown in \cref{tab:wi_ose_a}).
Furthermore, our method \textbf{does not} require exemplar replay when learning new classes, enabling end-to-end OWOD.
\begin{table*}[ht]
\centering
\scriptsize
\caption{\textbf{OWOD results on M-OWODB (top) and S-OWODB (bottom).} Our method largely outperforms the SOTA methods 
in terms of both known mAP and unknown recall (U-Recall) on both benchmarks. 
\ddag~is the unmodified YOLO-World detector~\cite{yolow} prompted with known class names and a hand-crafted generic object name (``object''). \dag~uses a pretrained language model to learn the semantic topology of classes. * denotes models that involve pretrained vision–language models. Other results are directly taken from \cite{sun2024exploring}.}
\setlength{\tabcolsep}{3pt}
\adjustbox{width=\textwidth}{
\begin{tabular}{@{}l|cc|cccc|cccc|ccc@{}}
\toprule
 \textbf{Task IDs} ($\rightarrow$)& \multicolumn{2}{c|}{\textbf{Task 1}} & \multicolumn{4}{c|}{\textbf{Task 2}} & \multicolumn{4}{c|}{\textbf{Task 3}} & \multicolumn{3}{c}{\textbf{Task 4}} \\ \midrule
 
 & \cellcolor[HTML]{FFFFED}{U-Recall} & \multicolumn{1}{c|}{\cellcolor[HTML]{EDF6FF}{mAP ($\uparrow$)}} & \cellcolor[HTML]{FFFFED}{U-Recall} & \multicolumn{3}{c|}{\cellcolor[HTML]{EDF6FF}{mAP ($\uparrow$)}} & \cellcolor[HTML]{FFFFED}{U-Recall} & \multicolumn{3}{c|}{\cellcolor[HTML]{EDF6FF}{mAP ($\uparrow$)}} & \multicolumn{3}{c}{\cellcolor[HTML]{EDF6FF}{mAP ($\uparrow$)}}  \\

Method & \cellcolor[HTML]{FFFFED}($\uparrow$) & \begin{tabular}[c]{@{}c}Current \\ known\end{tabular} & \cellcolor[HTML]{FFFFED}($\uparrow$) & \begin{tabular}[c]{@{}c@{}}Previously\\  known\end{tabular} & \begin{tabular}[c]{@{}c@{}}Current \\ known\end{tabular} & Both & \cellcolor[HTML]{FFFFED}($\uparrow$) & \begin{tabular}[c]{@{}c@{}}Previously \\ known\end{tabular} & \begin{tabular}[c]{@{}c@{}}Current \\ known\end{tabular} & Both & \begin{tabular}[c]{@{}c@{}}Previously \\ known\end{tabular} & \begin{tabular}[c]{@{}c@{}}Current \\ known\end{tabular} & Both \\ \midrule

ORE~\cite{TowardsOWOD} & \cellcolor[HTML]{FFFFED}4.9  & 56.0 & \cellcolor[HTML]{FFFFED}2.9 & 52.7 & 26.0 & 39.4  & \cellcolor[HTML]{FFFFED}3.9 & 38.2 & 12.7 & 29.7 & 29.6 & 12.4 & 25.3 \\ 

OST\dag~\cite{ost} & \cellcolor[HTML]{FFFFED}- & 56.2 & \cellcolor[HTML]{FFFFED}- & 53.4 & 26.5 & 39.9 & \cellcolor[HTML]{FFFFED}- & 38.0 & 12.8 & 29.6 & 30.1 & 13.3 & 25.9 \\

OW-DETR~\cite{OW-DETR} & \cellcolor[HTML]{FFFFED}7.5  & 59.2 & \cellcolor[HTML]{FFFFED}6.2 & 53.6 & 33.5 & 42.9 & \cellcolor[HTML]{FFFFED}5.7 & 38.3 & 15.8 & 30.8 & 31.4 & 17.1 & 27.8 \\

UC-OWOD~\cite{UC-OWOD} & \cellcolor[HTML]{FFFFED}2.4  & 50.7 & \cellcolor[HTML]{FFFFED}3.4 & 33.1 & 30.5 & 31.8  & \cellcolor[HTML]{FFFFED}8.7 & 28.8 & 16.3 & 24.6 & 25.6 & 15.9 & 23.2 \\ 

ALLOW~\cite{ALLOW} & \cellcolor[HTML]{FFFFED}13.6 & 59.3&\cellcolor[HTML]{FFFFED}10.0& 53.2 &34.0& 45.6&\cellcolor[HTML]{FFFFED}14.3& 42.6 &26.7& 38.0 &33.5 &21.8& 30.6 \\

PROB~\cite{prob}  & \cellcolor[HTML]{FFFFED}19.4  & 59.5 & \cellcolor[HTML]{FFFFED}17.4 & 55.7 & 32.2 & 44.0  & \cellcolor[HTML]{FFFFED}19.6 & 43.0 & 22.2 & 36.0 & 35.7 & 18.9 & 31.5 \\

CAT~\cite{cat} & \cellcolor[HTML]{FFFFED}23.7 & 60.0 &\cellcolor[HTML]{FFFFED}19.1 & 55.5 & 32.7 & 44.1 &\cellcolor[HTML]{FFFFED}24.4 & 42.8 & 18.7 & 34.8 & 34.4 & 16.6 & 29.9 \\

RandBox~\cite{randbox}  & \cellcolor[HTML]{FFFFED}10.6 & 61.8 & \cellcolor[HTML]{FFFFED}6.3 &- &- &45.3& \cellcolor[HTML]{FFFFED}7.8 & - &- &39.4 &- &- &35.4 \\

EO-OWOD~\cite{sun2024exploring}   & \cellcolor[HTML]{FFFFED}24.6 & 61.3 &\cellcolor[HTML]{FFFFED}26.3 & 55.5 & 38.5 & 47.0 &\cellcolor[HTML]{FFFFED}29.1 & 46.7 & 30.6 & 41.3 & 42.4 & 24.3 & 37.9 \\

\midrule

MAVL*~\cite{Maaz2022Multimodal} & \cellcolor[HTML]{FFFFED} 50.1 & 64.0 &\cellcolor[HTML]{FFFFED}49.5 & 61.6 & 30.8 & 46.2 &\cellcolor[HTML]{FFFFED}50.9 & 43.8 &22.7 & 36.8 & 36.2 & 20.6 & 32.3 \\

SKDF*~\cite{skdf} & \cellcolor[HTML]{FFFFED}39.0 & 56.8 & \cellcolor[HTML]{FFFFED}36.7 & 52.3 & 28.3 & 40.3 & \cellcolor[HTML]{FFFFED}36.1 & 36.9 & 16.4 & 30.1 & 31.0 & 14.7 & 26.9 \\

YOLO-World\ddag & \cellcolor[HTML]{FFFFED}16.6 & 71.9 & \cellcolor[HTML]{FFFFED}16.1 & 71.8 & 48.1 & 60.0 & \cellcolor[HTML]{FFFFED}13.0 & 60.0 & 40.7 & 53.6 & 53.7 & 33.9 & 48.7 \\
\textbf{Ours} & \cellcolor[HTML]{FFFFED}\textbf{73.5} & \textbf{72.1} & \cellcolor[HTML]{FFFFED}\textbf{77.5} & \textbf{72.4} & \textbf{51.0} & \textbf{61.7} & \cellcolor[HTML]{FFFFED}\textbf{76.1} & \textbf{61.6} & \textbf{41.6} & \textbf{54.9} & \textbf{56.0} & \textbf{34.3} & \textbf{50.6} \\

\midrule
\midrule 

ORE~\cite{TowardsOWOD} & \cellcolor[HTML]{FFFFED}1.5  & 61.4 & \cellcolor[HTML]{FFFFED}3.9 & 56.5 & 26.1 & 40.6  & \cellcolor[HTML]{FFFFED}3.6 & 38.7 & 23.7 & 33.7 & 33.6 & 26.3 & 31.8 \\

OW-DETR~\cite{OW-DETR} & \cellcolor[HTML]{FFFFED}5.7  & 71.5 & \cellcolor[HTML]{FFFFED}6.2 & 62.8 & 27.5 & 43.8 & \cellcolor[HTML]{FFFFED}6.9 & 45.2 & 24.9 & 38.5 & 38.2 & 28.1 & 33.1 \\

PROB~\cite{prob} & \cellcolor[HTML]{FFFFED}17.6  & 73.4 & \cellcolor[HTML]{FFFFED}22.3 & 66.3 & 36.0 & 50.4 & \cellcolor[HTML]{FFFFED}24.8 & 47.8 & 30.4 & 42.0 & 42.6 & 31.7 & 39.9 \\

CAT~\cite{cat} &\cellcolor[HTML]{FFFFED}24.0 & 74.2 &\cellcolor[HTML]{FFFFED}23.0 & 67.6 & 35.5 & 50.7 &\cellcolor[HTML]{FFFFED}24.6 & 51.2 & 32.6 & 45.0 & 45.4 & 35.1 & 42.8 \\

EO-OWOD~\cite{sun2024exploring}   & \cellcolor[HTML]{FFFFED}24.6 & 71.6 &\cellcolor[HTML]{FFFFED}27.9 & 64.0 & 39.9 & 51.3 & \cellcolor[HTML]{FFFFED}31.9 & 52.1 & 42.2 & 48.8 & 48.7 & 38.8 & 46.2 \\
\midrule
SKDF*~\cite{skdf} & \cellcolor[HTML]{FFFFED}60.9 & 69.4 & \cellcolor[HTML]{FFFFED}60.0 & 63.8 & 26.9 & 44.4 & \cellcolor[HTML]{FFFFED}58.6 & 46.2 & 28.0 & 40.1 & 41.8 & 29.6 & 38.7 \\
YOLO-World \ddag & \cellcolor[HTML]{FFFFED}29.0 & 75.6 & \cellcolor[HTML]{FFFFED}26.1 & \textbf{75.7} & 55.3 & 65.0 & \cellcolor[HTML]{FFFFED}26.9 & 65.1 & 54.4 & 61.6 & 61.4 & 55.2 & 59.9 \\
\textbf{Ours} &  \cellcolor[HTML]{FFFFED}\textbf{71.3} &\textbf{76.4} & \cellcolor[HTML]{FFFFED}\textbf{74.4} & 75.0 & \textbf{59.8} & \textbf{67.0} &  \cellcolor[HTML]{FFFFED}\textbf{74.6} & \textbf{67.0} & \textbf{53.8} & \textbf{62.6} & \textbf{65.5} & \textbf{56.9} & \textbf{63.4} \\ 

\bottomrule

\end{tabular}%
}\vspace{-0.75cm}
\label{table1}
\end{table*}

\begin{table*}[h]
\centering
\scriptsize
\caption{ \textbf{Unknown Object Confusion on M-OWODB.} Wilderness impact (WI) and absolute open set error (A-OSE) reflect the negative impact of unknown objects on the accuracy of known classes. These metrics do not apply to task 4 since all classes are known. 
}
\label{tab:wi_ose_a}
\setlength{\tabcolsep}{10pt}
\adjustbox{width=\textwidth}{
\begin{tabular}{@{}l|ccc|ccc|ccc}
\toprule
 \textbf{Task IDs} ($\rightarrow$)& \multicolumn{3}{c|}{\textbf{Task 1}} & \multicolumn{3}{c|}{\textbf{Task 2}} & \multicolumn{3}{c}{\textbf{Task 3}} \\
\midrule

& \cellcolor[HTML]{FFFFED}{U-Recall} & \cellcolor[HTML]{EDF6FF}{WI} & \cellcolor[HTML]{EDF6FF}{A-OSE} & \cellcolor[HTML]{FFFFED}{U-Recall} & \cellcolor[HTML]{EDF6FF}{WI} & \cellcolor[HTML]{EDF6FF}{A-OSE}  & \cellcolor[HTML]{FFFFED}{U-Recall} & \cellcolor[HTML]{EDF6FF}{WI} & \cellcolor[HTML]{EDF6FF}{A-OSE} \\

Method & \cellcolor[HTML]{FFFFED}($\uparrow$) & \cellcolor[HTML]{EDF6FF}($\downarrow$) & \cellcolor[HTML]{EDF6FF}($\downarrow$) & \cellcolor[HTML]{FFFFED}($\uparrow$) & \cellcolor[HTML]{EDF6FF}($\downarrow$) & \cellcolor[HTML]{EDF6FF}($\downarrow$) & \cellcolor[HTML]{FFFFED}($\uparrow$) & \cellcolor[HTML]{EDF6FF}($\downarrow$) & \cellcolor[HTML]{EDF6FF}($\downarrow$) \\

 \midrule



ORE~\cite{TowardsOWOD} & \cellcolor[HTML]{FFFFED}4.9  & 0.0621 & 10459 & \cellcolor[HTML]{FFFFED}2.9 & 0.0282 & 10445 & \cellcolor[HTML]{FFFFED}3.9 & 0.0211 & 7990  \\ 

OST\dag~\cite{ost}  &\cellcolor[HTML]{FFFFED}- & 0.0417 & 4889 &\cellcolor[HTML]{FFFFED}- & 0.0213 & 2546 &\cellcolor[HTML]{FFFFED}- & 0.0146 & 2120 \\

OW-DETR \cite{OW-DETR} & \cellcolor[HTML]{FFFFED}7.5  & 0.0571 & 10240 & \cellcolor[HTML]{FFFFED}6.2 & 0.0278 & 8441 & \cellcolor[HTML]{FFFFED}5.7 & 0.0156 & 6803  \\

PROB~\cite{prob} & \cellcolor[HTML]{FFFFED}19.4  & 0.0569 & 5195 & \cellcolor[HTML]{FFFFED}17.4 & 0.0344 & 6452 & \cellcolor[HTML]{FFFFED}19.6 & 0.0151 &2641  \\

RandBox~\cite{randbox} & \cellcolor[HTML]{FFFFED}10.6 & 0.0240 & 4498 & \cellcolor[HTML]{FFFFED}6.3 & 0.0078 & 1880 & \cellcolor[HTML]{FFFFED}7.8 & 0.0054 & 1452 \\

EO-OWOD~\cite{sun2024exploring} & \cellcolor[HTML]{FFFFED}24.6 & 0.0299 & 4148 & \cellcolor[HTML]{FFFFED}26.3 & 0.0099 & 1791 & \cellcolor[HTML]{FFFFED}29.1 & 0.0077 & 1345 \\

\midrule

YOLO-World & \cellcolor[HTML]{FFFFED}16.6  & 0.0311 & 9070 & \cellcolor[HTML]{FFFFED}16.1 & 0.0147 & 7063 & \cellcolor[HTML]{FFFFED}13.0 & 0.0086 & 5060 \\

\textbf{Ours} & \cellcolor[HTML]{FFFFED}\textbf{73.5} & \textbf{0.0175} & \textbf{1038} & \cellcolor[HTML]{FFFFED}\textbf{77.5} & \textbf{0.0047} & \textbf{529} & \cellcolor[HTML]{FFFFED}\textbf{76.1} & \textbf{0.0030} & \textbf{448} \\
\bottomrule

\end{tabular}%
}\vspace{-0.75cm}
\end{table*}

\begin{table*}[ht]
\centering
\scriptsize
\caption{\textbf{Evaluation on nu-OWODB.} Our method achieves leading performance in mAP for known classes and U-Recall for unknown classes on the benchmark based on real-world driving scenes. (ft) indicates a method that fine-tunes the model after learning new tasks with exemplars from the previous task, which is not applicable for task 1.} 

\setlength{\tabcolsep}{3pt}
\adjustbox{width=\textwidth}{
\begin{tabular}{@{}l|cccc|cccccc|cccc|ccc@{}}
\toprule
 \textbf{Task IDs} ($\rightarrow$)& \multicolumn{4}{c|}{\textbf{Task 1}} & \multicolumn{6}{c|}{\textbf{Task 2}} & \multicolumn{3}{c}{\textbf{Task 3}} \\ \midrule
 
 & \cellcolor[HTML]{FFFFED}{U-Recall} & \cellcolor[HTML]{EDF6FF}{WI} & \cellcolor[HTML]{EDF6FF}{A-OSE} & \multicolumn{1}{c|}{\cellcolor[HTML]{EDF6FF}{mAP ($\uparrow$)}} & \cellcolor[HTML]{FFFFED}{U-Recall} & \cellcolor[HTML]{EDF6FF}{WI} & \cellcolor[HTML]{EDF6FF}{A-OSE} & \multicolumn{3}{c|}{\cellcolor[HTML]{EDF6FF}{mAP ($\uparrow$)}}  & \multicolumn{3}{c}{\cellcolor[HTML]{EDF6FF}{mAP ($\uparrow$)}}  \\

Method & \cellcolor[HTML]{FFFFED}($\uparrow$) &  \cellcolor[HTML]{EDF6FF}($\downarrow$) & \cellcolor[HTML]{EDF6FF}($\downarrow$) & \begin{tabular}[c]{@{}c}Current \\ known\end{tabular} & \cellcolor[HTML]{FFFFED}($\uparrow$) & \cellcolor[HTML]{EDF6FF}($\downarrow$) & \cellcolor[HTML]{EDF6FF}($\downarrow$) & \begin{tabular}[c]{@{}c@{}}Previously\\  known\end{tabular} & \begin{tabular}[c]{@{}c@{}}Current \\ known\end{tabular} & Both  & \begin{tabular}[c]{@{}c@{}}Previously \\ known\end{tabular} & \begin{tabular}[c]{@{}c@{}}Current \\ known\end{tabular} & Both  \\ \midrule
PROB~\cite{prob} & \cellcolor[HTML]{FFFFED}0.5&\textbf{0.0025}&2897&25.1& \cellcolor[HTML]{FFFFED}2.4&\textbf{0.0007}&751&0.0&7.7&3.2&0.1&14.9&3.9\\
PROB~\cite{prob} (ft) & \cellcolor[HTML]{FFFFED}-&-&-&-& \cellcolor[HTML]{FFFFED}2.8&0.0015&1583&27.2&6.7&18.8&18.1&16.0&17.5\\
EO-OWOD~\cite{sun2024exploring} & \cellcolor[HTML]{FFFFED}1.4&0.0059&\textbf{223}&22.4& \cellcolor[HTML]{FFFFED}0.0&0.0017&\textbf{28}&0.0&9.6&3.9&0.0&24.5&6.4\\
EO-OWOD~\cite{sun2024exploring} (ft) & \cellcolor[HTML]{FFFFED}-&-&-&-& \cellcolor[HTML]{FFFFED}0.8&0.0030&172&27.0&13.5&21.4&21.8&\textbf{25.6}&22.8  \\

\midrule

YOLO-World  &  \cellcolor[HTML]{FFFFED}2.1 & 0.0463 & 12316 & 21.8 &  \cellcolor[HTML]{FFFFED}3.2 & 0.0141 & 4486 & 21.8 & 5.1 & 14.9 & 14.8 & 9.3 & 13.4 \\
\textbf{Ours} &  \cellcolor[HTML]{FFFFED}\textbf{45.5} & 0.0185 & 1724 & \textbf{28.1} &  \cellcolor[HTML]{FFFFED}\textbf{40.8} & 0.0106 & 1703 & \textbf{27.8} & \textbf{15.5} & \textbf{22.8} & \textbf{23.8} & 25.3 & \textbf{24.2}\\

\bottomrule

\end{tabular}%
}\vspace{-0.75cm}
\label{table1_nu}
\end{table*}

\paragraph{Quantitative Results on Driving Scenes} We further evaluate our method and some SOTA methods on nu-OWODB, a new benchmark based on nuScenes~\cite{nuscenes2019}. 
As shown in \cref{table1_nu}, our method achieves a clear advantage in U-Recall across all tasks, surpassing state-of-the-art performance by up to 40\% — despite the significant domain gap between nuScenes and vision-language pretraining datasets, as evidenced by the low mAP and U-Recall scores of the unmodified YOLO-World baseline. Our method also has the highest mAP for known classes in each task.
Although our method does not need any re-training on known classes, we still allow existing methods~\cite{prob,sun2024exploring} to fine-tune the model with exemplars from the previous task (10\% data) after learning new task, otherwise they will exhibit catastrophic forgetting. As a result, they achieve better WI and A-OSE compared to our method. To some extent, these metrics can reflect the robustness of object detectors, but they can also be misleading considering that the models make very few predictions (see Task 2).

We also observe that the mAP for known classes is low. The main reason is that object categories are highly detailed and objects in each class are highly imbalanced in nuScenes~\cite{nuscenes2019}. For example, the class vehicle.bus.rigid has 8361 2D annotations, while the class vehicle.bus.bendy has only 265 annotations. We did not merge these categories, to make the benchmark more realistic and challenging.

\begin{figure}[h]
\begin{minipage}[b]{0.48\textwidth}
\textbf{Quantitative Results on OVD benchmark}
Our method performs OWOD by optimizing class embeddings and additional MSCAL modules, while keeping the parameters adapted from YOLO-World frozen. As a result, our model maintains the performance of open-vocabulary object detection in a zero-shot manner.
\end{minipage}\hfill
\begin{minipage}[b]{0.48\linewidth}
\captionof{table}{\textbf{Zero-shot open vocabulary performance on LVIS minival.}}
\centering
\scriptsize
\begin{tabular}{lcccc}
\toprule
\textbf{Model} & \textbf{$AP$} & \textbf{$AP_r$} & \textbf{$AP_c$} & \textbf{$AP_f$} \\
\midrule
MDETR~\cite{goldg} & 24.2 & 20.9 & 24.3 & 24.2 \\
Grounding DINO-T~\cite{groundingdino} & 27.4 & 18.1 & 23.3 & 32.7 \\
DetCLIP-T~\cite{DetCLIP} & 34.4 & 26.9 & 33.9 & 36.3 \\
YOLO-World-XL~\cite{yolow} & 35.7 & 26.4 & 33.9 & 39.0 \\
Ours  & 35.7 & 26.4 & 33.9 & 38.9 \\
\bottomrule
\end{tabular}
\label{lvis-zs}
\end{minipage}
\vspace{-0.3cm}
\end{figure}

Following~\cite{yolow}, we evaluate the zero-shot performance on LVIS minival~\cite{goldg} (1203 classes). Known Class Embeddings are initialized from class names, while the Pseudo Unknown Embedding is constructed from class names and the ``object'' prompt. \cref{lvis-zs} shows performance comparable to state-of-the-art OVD methods, confirming that our framework unifies OVD and OWOD tasks.

\begin{figure*}[ht]
  \centering
    \includegraphics[width=0.95\textwidth]{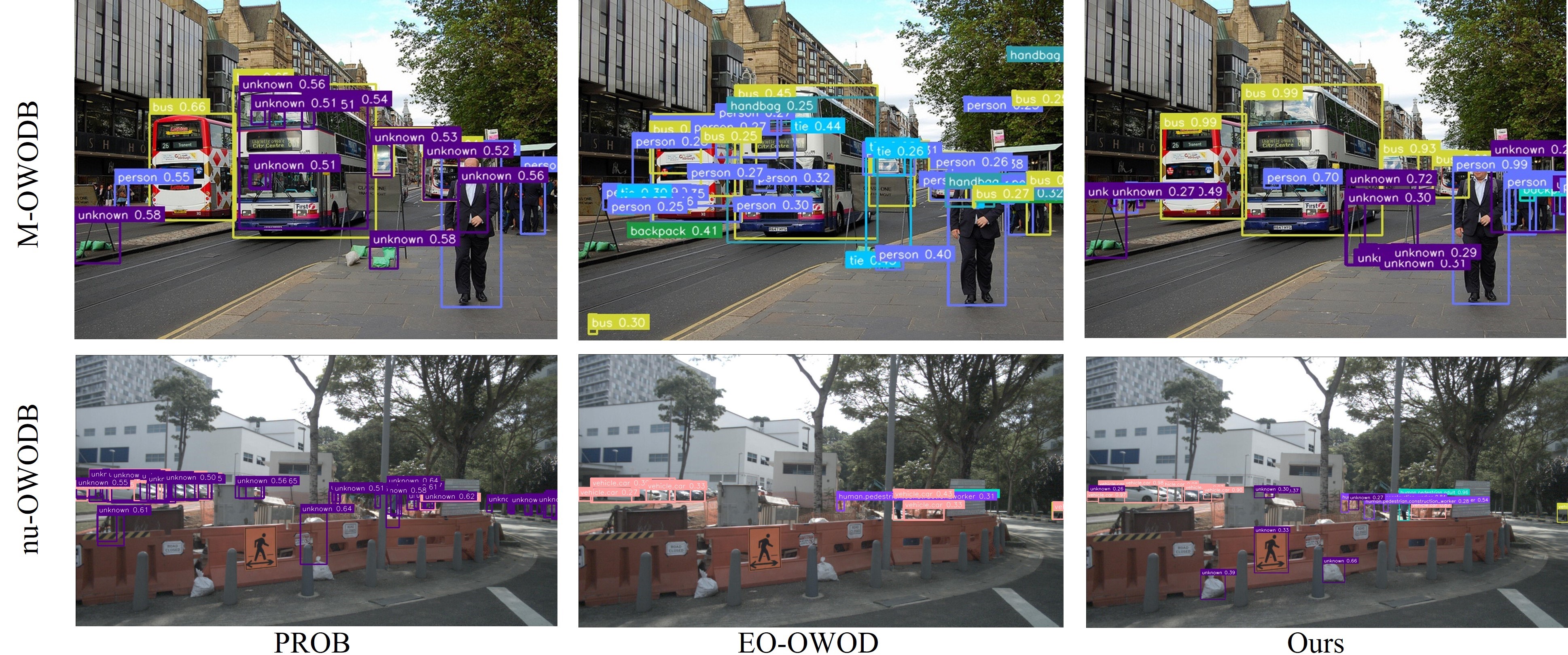}
    \caption{\textbf{Qualitative results on M-OWODB and nu-OWODB.} Our method produces bounding boxes of known and unknown objects with better quality compared to PROB~\cite{prob} and EO-OWODB~\cite{sun2024exploring}.} 
    \label{fig:vis}
\end{figure*}

\paragraph{Qualitative Results} \cref{fig:vis} shows a comparison between two SOTA methods and our method after completing the second task of M-OWODB and nu-OWODB. Our method provides more meaningful bounding boxes for unknown classes, without significantly losing performance on known classes, which makes it easier to make trade-off between precision and recall in various applications.
We can see that although PROB~\cite{prob} is able to detect a reasonable number of unknown objects, but there are also many high-confidence unknown bounding boxes associated with known objects. On the other hand, EO-OWODB~\cite{sun2024exploring} detects background as known objects while failing to detect several unknown objects.

\begin{table*}[ht]
\centering
\caption{\textbf{Ablation study on nu-OWODB.} The comparison is shown in terms of mean average precision (mAP), wilderness impact (WI), absolute open set error (A-OSE) and unknown class recall (U-Recall).}
\setlength{\tabcolsep}{3pt}
\adjustbox{width=\textwidth}{
\begin{tabular}{@{}l|cccc|cccccc|cccc|ccc@{}}
\toprule
 \textbf{Task IDs} ($\rightarrow$)& \multicolumn{4}{c|}{\textbf{Task 1}} & \multicolumn{6}{c|}{\textbf{Task 2}} & \multicolumn{3}{c}{\textbf{Task 3}} \\ \midrule
 
 & \cellcolor[HTML]{FFFFED}{U-Recall} & \cellcolor[HTML]{EDF6FF}{WI} & \cellcolor[HTML]{EDF6FF}{A-OSE} & \multicolumn{1}{c|}{\cellcolor[HTML]{EDF6FF}{mAP ($\uparrow$)}} & \cellcolor[HTML]{FFFFED}{U-Recall} & \cellcolor[HTML]{EDF6FF}{WI} & \cellcolor[HTML]{EDF6FF}{A-OSE} & \multicolumn{3}{c|}{\cellcolor[HTML]{EDF6FF}{mAP ($\uparrow$)}}  & \multicolumn{3}{c}{\cellcolor[HTML]{EDF6FF}{mAP ($\uparrow$)}}  \\

Method & \cellcolor[HTML]{FFFFED}($\uparrow$) &  \cellcolor[HTML]{EDF6FF}($\downarrow$) & \cellcolor[HTML]{EDF6FF}($\downarrow$) & \begin{tabular}[c]{@{}c}Current \\ known\end{tabular} & \cellcolor[HTML]{FFFFED}($\uparrow$) & \cellcolor[HTML]{EDF6FF}($\downarrow$) & \cellcolor[HTML]{EDF6FF}($\downarrow$) & \begin{tabular}[c]{@{}c@{}}Previously\\  known\end{tabular} & \begin{tabular}[c]{@{}c@{}}Current \\ known\end{tabular} & Both  & \begin{tabular}[c]{@{}c@{}}Previously \\ known\end{tabular} & \begin{tabular}[c]{@{}c@{}}Current \\ known\end{tabular} & Both  \\ \midrule

Base Model  & \cellcolor[HTML]{FFFFED}2.1 & 0.0463 & 12316 & 21.8 & \cellcolor[HTML]{FFFFED}3.2 & 0.0141 & 4486 & 21.8 & 5.1 & 14.9 & 14.8 & 9.3 & 13.4 \\
OWEL & \cellcolor[HTML]{FFFFED}24.5 & 0.0381 & 18241 & 30.0 & \cellcolor[HTML]{FFFFED}24.0 & 0.0106 & 8827 & 29.5 & 15.9 & 23.9 & 23.8 & 25.3 & 24.2 \\
MSCAL & \cellcolor[HTML]{FFFFED}28.8 & 0.0178 & 1653 & 28.2 & \cellcolor[HTML]{FFFFED}24.7 & 0.0113 & 1772 & 27.9 & 15.8 & 23.0 & 23.8 & 25.3 & 24.2\\
\textbf{Ours} & \cellcolor[HTML]{FFFFED}45.5 & 0.0185 & 1724 & 28.1 & \cellcolor[HTML]{FFFFED}40.8 & 0.0106 & 1703 & 27.8 & 15.5 & 22.8 & 23.8 & 25.3 & 24.2\\

\bottomrule

\end{tabular}%
}\vspace{-0.75cm}
\label{ablation}
\end{table*}

\subsection{Ablation Study}

To understand the contribution of individual components, we disable some modules of our model to create a set of baseline models. Base Model is the vanilla YOLO-World detector~\cite{yolow} prompted with known class names and a hand-crafted
generic object name (“object”), in a zero shot manner. OWEL removes MSCAL modules and OOD scores, while the MSCAL replaces the pseudo unknown embedding $w_{\mathcal{U}}$ with the original generic prompt $w_0$.

From \cref{ablation}, we can see that OWEL significantly improves the U-Recall for unknown classes and mAP of known classes.  On the other hand, OWEL increases absolute open set error. This indicates that optimizing known class embeddings with object detection loss not only learns embeddings most similar to image samples, but also learns some characteristics of this domain, which leads to more valid predictions and open set errors. 
MSCAL largely reduces the open set error, and achieves reasonable unknown recall without pseudo unknown embedding, because it detects unknown classes by correcting the open set error. OWEL and MSCAL complement each other to detect the largest proportion of unknown objects while reasonably maintaining the performance on known classes.

\section{Conclusion} 
In this work, we propose a framework that enables open vocabulary object detectors to operate in open world settings without compromising their zero-shot capabilities. We introduce Open World Embedding Learning (OWEL) and Multi-Scale Contrastive Anchor Learning (MSCAL) which enable the model to identify and incrementally learn unknown objects. We further propose a new benchmark to evaluate the performance of OWOD for autonomous driving. In the future, we will study open world object detection with various sensor modalities and data domains.

\section*{Acknowledgments}
\noindent The first two authors acknowledge the financial support from The University of Melbourne through the Melbourne Research Scholarship. This research was supported by The University of Melbourne’s Research Computing Services and the Petascale Campus Initiative.

\bibliography{egbib}

\subsection{Generic Objectness Prompt}

\begin{table}[htbp]
\centering
\scriptsize
\caption{\textbf{Ablation study of different prompts} on M-OWODB task 1, where 20 classes in PASCAL VOC are known.}
\begin{tabular}{lcccc}
\toprule
\textbf{Generic Prompt} & \textbf{mAP $\uparrow$} & \textbf{U-Recall $\uparrow$} & \textbf{WI $\downarrow$} & \textbf{A-OSE $\downarrow$} \\
\midrule
\texttt{object}  & 71.9 & 16.6 & 0.0311 & 9070 \\
\texttt{entity} & 71.9 & 0.9 & 0.0353 & 10883 \\
\texttt{unknown} & 71.9 & 1.6 & 0.0361 & 10743 \\
\texttt{anything} & 71.9 & 5.9 & 0.0326 & 10154 \\
\texttt{everything} & 71.9 & 4.3 & 0.0334 & 10379 \\
\bottomrule
\end{tabular}
\label{handcraft}
\end{table}

We tried different prompts to estimate the embedding for generic objectness. \cref{handcraft} shows the ablation experiment on M-OWODB, which shows the word ``object'' is an appropriate. 
We define ``object'' as the generic prompt in one experiment, then reuse it in all OWOD benchmarks (M-OWODB, S-OWODB, and nu-OWODB) without hand-crafted selection of the prompt across domains.
Though it is possible to estimate the generic prompt by designing a pretext task, but this will introduce bias towards known classes.

\subsection{Fine-tuning YOLO-World}

\begin{table}[htbp]
\caption{Open world performance on nu-OWODB task 1.}
\centering
\scriptsize
\begin{tabular}{lcccc}
\toprule
\textbf{Method} & \textbf{mAP $\uparrow$} & \textbf{U-Recall $\uparrow$} & \textbf{WI $\downarrow$} & \textbf{A-OSE $\downarrow$} \\
\midrule
YOLO-World (zero-shot) & 21.8 & 2.1 & 0.0463 & 12316  \\
YOLO-World (Fine-tuned) &  \textbf{30.0} & 4.9 & 0.0419 & 20039  \\
Ours &  28.1 & \textbf{45.5} & \textbf{0.0185} & \textbf{1724} \\
\bottomrule
\end{tabular}
\label{yolowft1}
\end{table}

Although OVD models perform zero-shot detection by design, we can still fine-tune the model for better performance, especially on datasets from different domains. However, this introduces additional problems in the open-world scenario. As shown in~\cref{yolowft1}, when we fine-tune the class embeddings of YOLO-World on nu-OWODB, the closed-set performance (mAP) improves, but the model's resistance to OOD objects is significantly reduced, resulting in a higher A-OSE. In contrast, our method shows significant gains in both known and unknown performance, highlighting the importance of the proposed Pseudo Unknown Embedding and Multi-Scale Contrastive Anchor Learning.

\subsection{Ablation on \texorpdfstring{$\alpha$}{alpha}}
As shown in \cref{alpha_eq2}, we try different $\alpha$ value varing from 0.2 to 0.8. The result shows that the choice of $\alpha$ does not make a significant impact on performance.

\begin{table}[htbp]
\centering
\small
\caption{\textbf{Ablation study of $\alpha$ on nu-OWODB task 1.}}
\begin{tabular}{lcccc}
\toprule
\textbf{$\alpha$} & \textbf{mAP $\uparrow$} & \textbf{U-Recall $\uparrow$} & \textbf{WI $\downarrow$} & \textbf{A-OSE $\downarrow$} \\
\midrule
0.2  & 28.2 & 38.2 & 0.0183 & 1700 \\
0.4 &  28.1 & 45.5 & 0.0185 & 1724 \\
0.8 & 27.6 & 37.8 & 0.0166 & 1455 \\
\bottomrule
\end{tabular}
\label{alpha_eq2}
\end{table}

\section{Discussion on Evaluation Metrics}
In this work, we use 4 commonly used evaluation metrics.
For known classes, the commonly used metric is mean average precision (mAP). 
For unknown classes, the primary metric is unknown class recall (U-Recall), which assesses the ratio of detected unknown objects. In addition, we choose wilderness impact (WI) \cite{elephant} and absolute open-set error (A-OSE) \cite{miller18} as secondary evaluation metrics. 

U-Recall and mAP are intuitive and widely adopted evaluation metrics in OWOD.
A-OSE and WI are primarily designed for open set object detection, and evaluate the interference of unknown objects on the detection performance of known objects.
A-OSE shows the absolute number of unknown objects detected as known classes at 0.5 IoU threshold. WI measures the impact of unknown objects on the model's precision. The definition of WI is:

\begin{equation}
WI = \frac{P_{\mathcal{K}}}{P_{\mathcal{K}\cup\mathcal{U}}} - 1,
\end{equation}

where $P_{\mathcal{K}}$ is the precision in closed set and $P_{\mathcal{K}\cup\mathcal{U}}$ is the precision in open set. Following \cite{TowardsOWOD}, we use the 0.8 recall threshold and 0.5 IoU threshold when calculating WI.

Although A-OSE and WI can somehow reflect the model's confusion between known and unknown objects, some limitations remain. In Table 2 (main paper), we can see that PROB~\cite{prob} and EO-WOOD~\cite{sun2024exploring} experience catastrophic forgetting after learning the new class, but they achieve the best A-OSE and WI. As a more extreme example, if the model does not output any bounding box at all, A-OSE will be 0. As a result, A-OSE and WI are only meaningful when models have similar precision.

\begin{figure*}[ht]
  \centering
    \includegraphics[width=\textwidth]{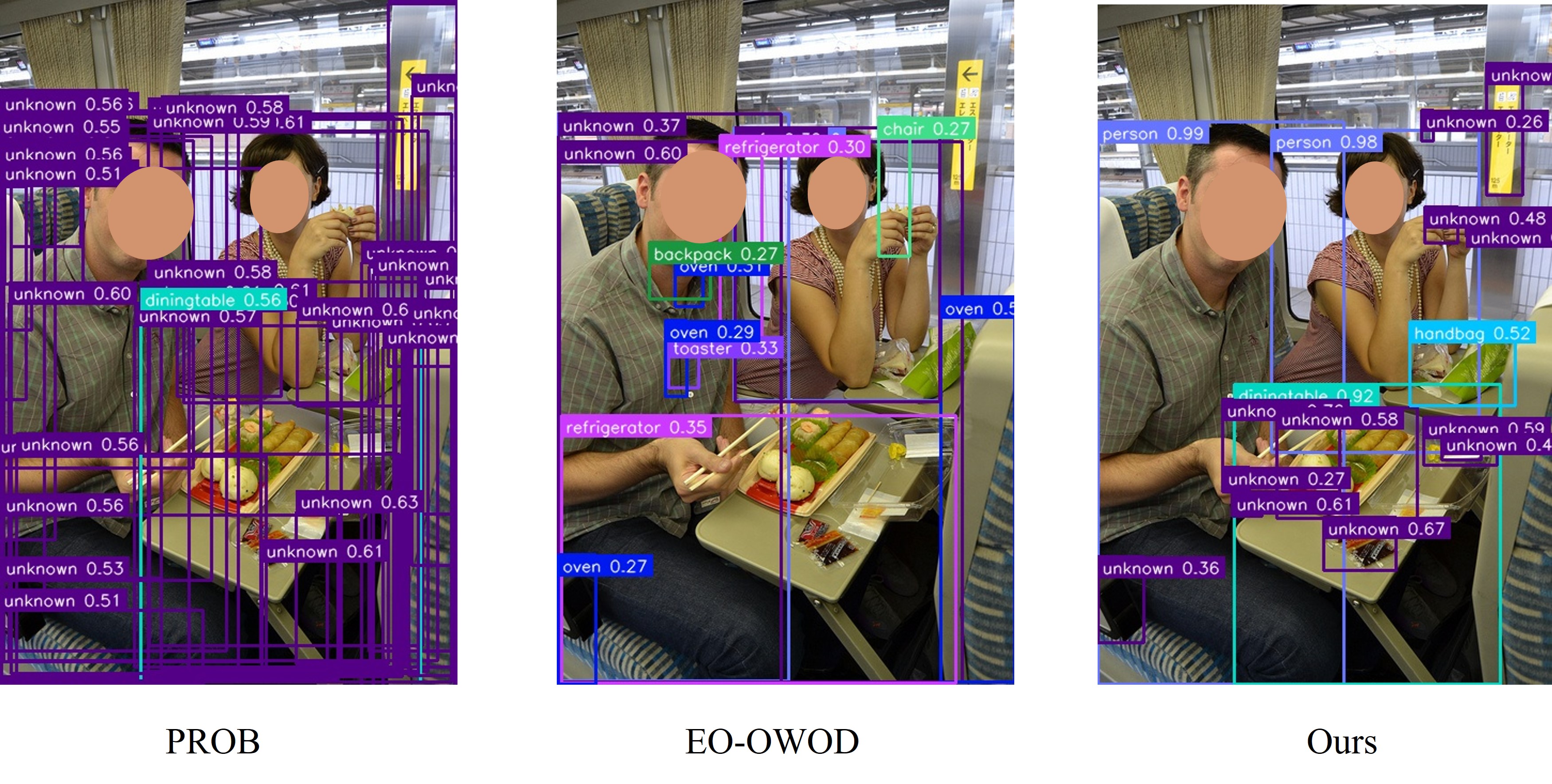}
    \caption{\textbf{Visualization of indoor scenes on M-OWODB.} We compare our method with PROB~\cite{prob} and EO-OWOD~\cite{sun2024exploring} using their official checkpoints on M-OWODB task2. Face occlusions are added after model inference.}
    \label{fig:m_owodb_indoor}
\end{figure*}

\begin{figure}[ht]
  \centering
    \includegraphics[width=0.9\textwidth]{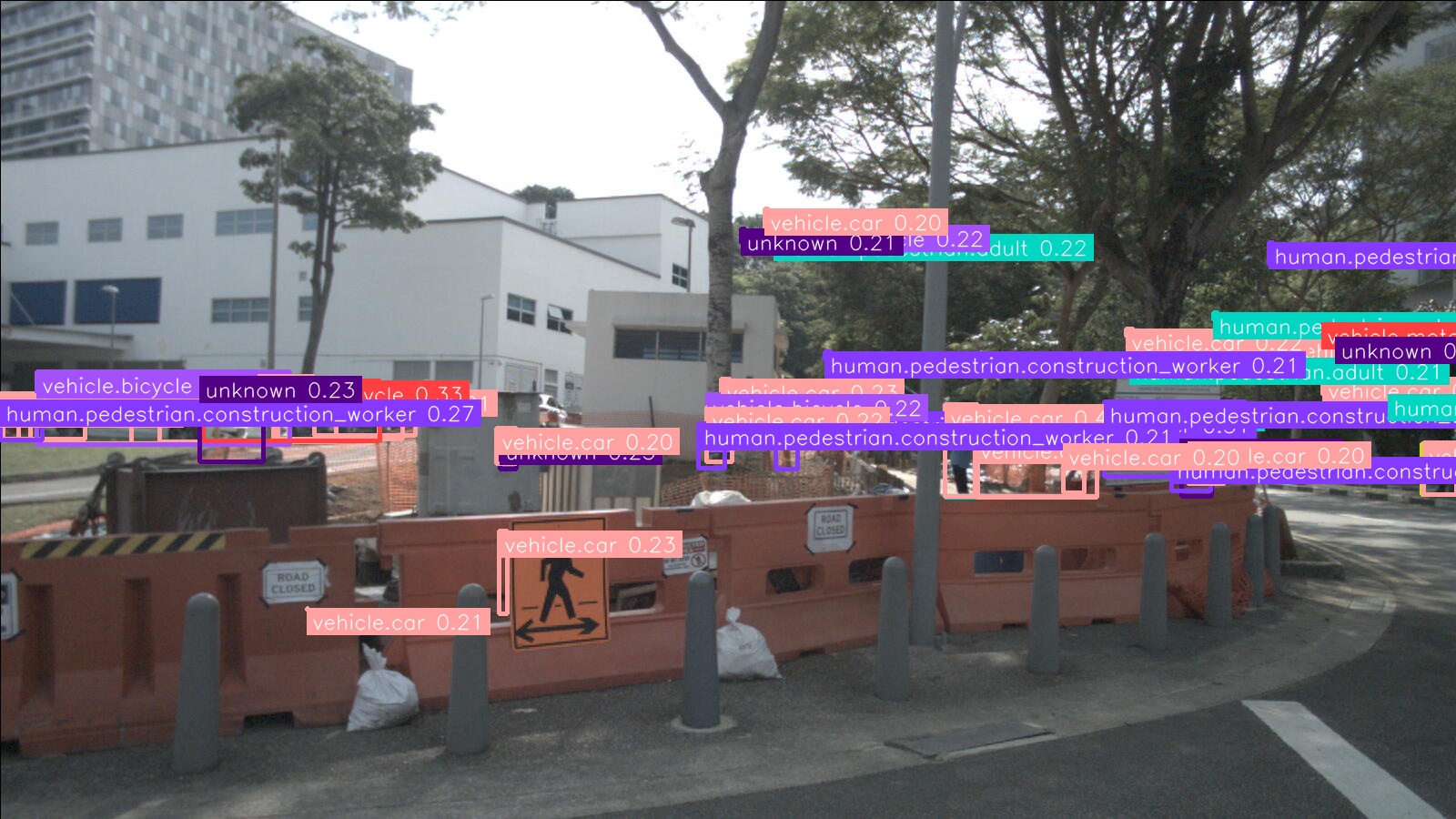}
    \caption{\textbf{Reducing the threshold leads to a significant drop in precision.} When EO-OWODB uses a threshold of 0.2, the number of bounding boxes of unknown classes increases slightly but also leads to many invalid bounding boxes.}
   \label{fig:eo302}
\end{figure}

\section{Additional Qualitative Results}
In \cref{fig:m_owodb_indoor}, we can see that the detection confidence is not always meaningful in OWOD because bounding boxes with high confidence can also be invalid. Monotonically scaling up the confidence score does not change the detection performance.
For all visualizations, we set the score threshold based on the principle of showing as many bounding boxes of unknown classes as possible without significantly reducing the precision of known classes. As shown in \cref{fig:eo302}, a lower score threshold leads to higher recall for EO-OWODB, but it also leads to a collapse in precision.
Although it is possible to set a threshold for each class separately, this will greatly reduce the ease of use of the model. The ease of use is ignored in evaluation metrics, but it is quite important for real-world applications.

\section{Details of nu-OWODB}
In this research, we present the nuScenes Open World Object Detection Benchmark (nu-OWODB), a novel benchmark designed to simulate the challenges of open-world object detection (OWOD) encountered in real world. Built on the nuImages subset of nuScenes~\cite{nuscenes2019}, the benchmark encompasses 23 highly diverse and imbalanced object classes. The dataset is publicly available at \href{https://www.nuscenes.org/nuimages}{www.nuscenes.org/nuimages}.

We divide the classes into three major tasks: vehicles, pedestrians and obstacles. The task-category mapping relationship is shown in \cref{tab: nu-OWOD}, and information about each task is provided in \cref{composition}. 


In addition, the naming convention in nuScenes is different from the natural language names of the classes. To fully utilize YOLO-World's zero shot capabilities, we define a natural language prompt for each class as shown in \cref{nuprompt}.
 
\begin{table}[htbp]
    \scriptsize
    \centering
    \caption{\textbf{Task-Category Mapping in nu-OWODB.}}
{%
    \begin{tabular}{l l}
    \toprule
    \textbf{Task}                     & \textbf{nuScenes Category}                            \\ \midrule
    Task 1 - Vehicles         & vehicle.bicycle \\
                              & vehicle.motorcycle \\
                              & vehicle.car \\
                              & vehicle.bus.bendy \\
                              & vehicle.bus.rigid \\
                              & vehicle.truck \\
                              & vehicle.emergency.ambulance \\
                              & vehicle.emergency.police \\
                              & vehicle.construction \\
                              & vehicle.trailer \\ \midrule
    Task 2 - Pedestrians      & human.pedestrian.adult \\
                              & human.pedestrian.child \\
                              & human.pedestrian.wheelchair \\
                              & human.pedestrian.stroller \\
                              & human.pedestrian.personal\_mobility \\
                              & human.pedestrian.police\_officer \\
                              & human.pedestrian.construction\_worker \\ \midrule
    Task 3 - Obstacles        & movable\_object.barrier \\
                              & movable\_object.trafficcone \\
                              & movable\_object.pushable\_pullable \\
                              & movable\_object.debris \\
                              & static\_object.bicycle\_rack \\
                              & animal \\ \bottomrule
    \end{tabular}%
    }
    \label{tab: nu-OWOD}
\end{table}

\begin{table}[htbp]
\scriptsize
\centering
\caption{\textbf{Task composition in nu-OWODB.} The semantics of each task split and the number of associated training and test images and object instances are displayed.}
\begin{tabular}{lccc}
\toprule
\textbf{Task IDs (→)} & \textbf{Task 1} & \textbf{Task 2} & \textbf{Task 3} \\
\midrule
 nu-OWODB & Vehicles & Pedestrians & Obstacles \\
\midrule
\# classes & 10 & 7 & 6 \\ 
\# training images & 53850 & 34957 & 25682 \\ 
\# test images & 13099 & 8473 & 6500 \\ 
\# training instances & 274587 & 135870 & 147253 \\ 
\# test instances & 64303 & 32710 & 39060 \\
\bottomrule
\end{tabular}
\label{composition}
\end{table}

\begin{table}[htbp]
    \scriptsize
    \centering
    \caption{\textbf{Correspondence between class names and text prompts in nu-OWODB.}}
\begin{tabular}{ll}
\toprule
\textbf{Class Name} & \textbf{Text Prompt} \\
\midrule
vehicle.bicycle & bicycle \\
vehicle.motorcycle & motorcycle \\
vehicle.car & car \\
vehicle.bus.bendy & articulated bus \\
vehicle.bus.rigid & rigid bus \\
vehicle.truck & truck \\
vehicle.emergency.ambulance & ambulance \\
vehicle.emergency.police & police car \\
vehicle.construction & construction vehicle \\
vehicle.trailer & trailer \\
human.pedestrian.adult & adult \\
human.pedestrian.child & child \\
human.pedestrian.wheelchair & wheelchair \\
human.pedestrian.stroller & stroller \\
human.pedestrian.personal\_mobility & scooter \\
human.pedestrian.police\_officer & police officer \\
human.pedestrian.construction\_worker & construction worker \\
movable\_object.barrier & barrier \\
movable\_object.trafficcone & traffic cone \\
movable\_object.pushable\_pullable & pushable and pullable object \\
movable\_object.debris & debris \\
static\_object.bicycle\_rack & bicycle rack \\
animal & animal \\
\bottomrule
\end{tabular}
\label{nuprompt}
\end{table}

\section{Implementation Details}
For the OVD component of our model, we adopt the general architecture of YOLO-World \cite{yolow}, which is pretrained on O365 \cite{o365}, GoldG \cite{goldg}, CC3M \cite{cc3m}. The backbone remains frozen in downstream tasks.
Importantly, images overlapped with COCO dataset \cite{coco} has been removed from GoldG by default \cite{yolow}.
Our model is trained using the AdamW optimizer \cite{adamw} with a learning rate of 1e-4 and a weight decay of 0.0125. The training batch size is set to 16, and the input images are rescaled to a resolution of 640×640. The weight parameter $\alpha$ is 0.4 for all tasks. All model training is conducted on a single Nvidia A100 GPU.

\section{Limitations and Future Work}
Although our method achieves state-of-the-art performance in various benchmarks, some limitations still exist. For example, we assume object embeddings cluster around the word ``object'' in the semantic space, but there can also be outliers. Alternatively, we may construct more pseudo unknown embeddings, but this will increase computational overhead.
For applications such as autonomous driving, we can further explore the use of LiDAR to detect class-agnostic objects and use the geometric information to perform 3D object detection in future work.

\end{document}